\documentclass[10pt,twocolumn,letterpaper]{article}

\usepackage[pagenumbers]{cvpr} % To force page numbers, e.g. for an arXiv version

\usepackage{graphicx}
\usepackage{amsmath}
\usepackage{amssymb}
\usepackage{booktabs}
\usepackage{makecell}

\usepackage[pagebackref,breaklinks,colorlinks]{hyperref}

\usepackage[capitalize]{cleveref}
\crefname{section}{Sec.}{Secs.}
\Crefname{section}{Section}{Sections}
\Crefname{table}{Table}{Tables}
\crefname{table}{Tab.}{Tabs.}

\def\cvprPaperID{1713} % *** Enter the CVPR Paper ID here
\def\confName{CVPR}
\def\confYear{2023}

\begin{document}

%%%%%%%%% TITLE - PLEASE UPDATE
\title{ITstyler: Image-optimized Text-based Style Transfer}

\author{%
  Yunpeng Bai$^{1}$, Jiayue Liu$^{1}$, Chao Dong$^{2,3}$, Chun Yuan$^{1,4}$ \\[0.5em]
  $^{1}$ Tsinghua University, $^{2}$Shenzhen Institutes of Advanced Technology, Chinese Academy of Sciences\\ $^{3}$ Shanghai AI Laboratory, China, $^{4}$Peng Cheng Laboratory, Shenzhen, China \\[0.3em]
%   {\small \texttt{\{chenh, bohe, hywang66,  yxren, abhinav\}@umd.edu, sernamlim@fb.com} }
}
\twocolumn[{%
\renewcommand\twocolumn[1][]{#1}%
\maketitle

\begin{center}
\centering
\includegraphics[width=0.97\textwidth,height=0.65\textwidth]{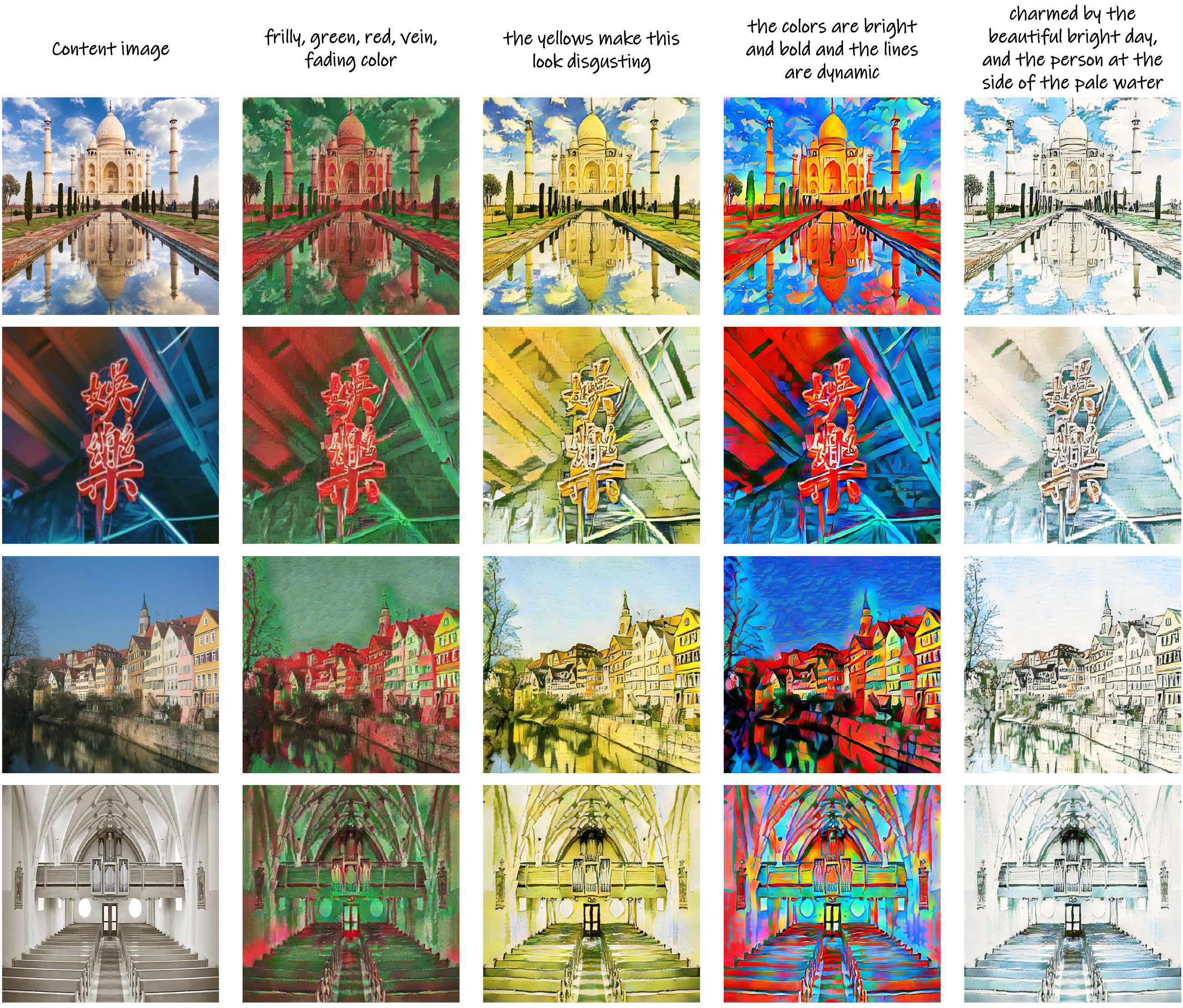}
 \vspace*{-0.2cm}
\captionof{figure}{Overview stylization results of the proposed method. These results are all obtained through one model, and the results are highly artistic while matching the text description well.}
\label{fig:overall}
\end{center}}]

% \begin{figure*}[htb]
%     \centering
%     \includegraphics[width=\linewidth]{figures/overall.pdf}
%     \caption{ Overall. }
%     \label{fig:overall}   
%     % \vspace{-0.1in}
% \end{figure*}

%%%%%%%%% ABSTRACT
\begin{abstract}
Text-based style transfer is a newly-emerging research topic that uses text information instead of style image to guide the transfer process, significantly extending the application scenario of style transfer. However, previous methods require extra time for optimization or text-image paired data, leading to limited effectiveness. In this work, we achieve a data-efficient text-based style transfer method that does not require optimization at the inference stage. Specifically, we convert text input to the style space of the pre-trained VGG network to realize a more effective style swap. We also leverage CLIP's multi-modal embedding space to learn the text-to-style mapping with the image dataset only. Our method can transfer arbitrary new styles of text input in real-time and synthesize high-quality artistic images.

% and propose a new effective text-based style transfer method called ITstyler.

% However, these methods do not match the image style with the text style well because the style representations they used are inappropriate. To solve this problem, we utilize the style feature space used in the previous image-based methods and convert the text input into representations that can describe styles well. 

%最近的一些工作研究了基于文本的图像风格化方法.用文本来作为一个风格的参考来来替代之前基于图像的方法中的风格图像，拓展了风格迁移的应用场景.然而这些方法并不能够很好地将图像的风格与文本的风格相匹配，因为他们对风格的表示方法有问题。
%我们利用了之前基于图像的方法所使用的的特征空间，将文本的转换为一种更加有效的风格表示。
%clip的encoder也被用来有效地抽取风格的特征，我们提出了一种新的有效的风格迁移方法，大量的结果表明了我们的方盒合成的结果非常好
%我们将文本的特征转换到一个适合做风格交换的空间，我们的方法以更加有效的方式将

% 

\end{abstract}

%%%%%%%%% BODY TEXT
\section{Introduction}
\label{sec:intro}

Artistic style transfer \cite{DBLP:conf/cvpr/GatysEB16,DBLP:conf/nips/LiFYWLY17,DBLP:conf/cvpr/ParkL19,chen2021artistic,li2019learning,sheng2018avatar} is an appealing research topic that aims to synthesize artworks by transferring artistic style from reference image to content image. The feature statistics from the pre-trained convolutional neural network were found able to describe the images' style, making it possible to achieve the neural style transfer. Extensive works \cite{DBLP:conf/cvpr/GatysEB16,DBLP:conf/iccv/HuangB17} have been studied to match the statistical information of two images and obtained amazing synthetic artistic results.

These image-based methods are already capable of producing nice artistic images, but all require a reference image as the style input. Therefore, if users want to create images of a specific style, they need to take the time to find a suitable style image first. Moreover, the style is independent of the content of the image. When providing style information, the content part of the style image is unnecessary and even brings interference to the results.

In order to extend the application scenario of style transfer, recent works have studied text-based style transfer. Instead of  finding a suitable style image, text descriptions can describe abstract styles more directly.
How to match the style features in different modalities of data is a challenge in text-based style transfer. CLIPstyler \cite {Kwon_2022_CVPR} utilizes the Contrastive Language-Image Pre-training (CLIP) \cite{radford2021learning} models and proposes a directional CLIP loss to align the CLIP-space direction between the text-image pairs of source and stylized result. However, for each text input, CLIPstyler needs to retrain the model, which is inefficient and unpractical. LDAST \cite{fu2022LDAST} is another recent work that uses two separate encoders to extract style and content features from text and image data. Then, an attention operation is used to fuse the two features to obtain stylized results. However, the operation is difficult to construct an accurate correspondence between features from different modalities. Besides, training their model also requires additional text-image paired data, which is usually difficult to obtain and hard to guarantee that text labels are matched with images. For these reasons, LDAST produces images of unsatisfactory quality.

These methods mentioned above do not find a suitable space for style transfer, resulting in limited effectiveness of image synthesis. Previous image-based methods do style transfer in the feature space of a pre-trained VGG \cite{DBLP:journals/corr/SimonyanZ14a} network, which has been proven capable of representing styles. We attempt to map the style described by the text to the image feature space and perform a more effective text-based style transfer.

In this work, we use the mean and variance of VGG features to represent style as in AdaIN \cite{DBLP:conf/iccv/HuangB17}. By combining this representation and CLIP's powerful capability for representing multi-modal features, we propose a new approach called ITstyler. We use the CLIP's text encoder to get style features and convert them to the corresponding style representation as the style input. With the help of CLIP's joint language-vision embedding space, we only use image data to train the network. Then it can also receive the text features as the style input. This elegant design makes ITstyler independent of text-image paired training data. The ability of VGG features to represent styles also improves the flexibility and performance of text-based style transfer.
% The ability of VGG features to represent styles also allows our approach to transfer arbitrary styles from text input and obtain the result in real-time.
%  using different data modalities 

% The most appealing property of ITstyler is that the training/testing phases use different data modalities. It adopts the image data for training but accepts the text data for testing. This elegant design makes ITstyler independent of test-image paired training data. It also improves the flexibility and performance of text-based style transfer.

To summarize, the main contributions of this work are as follows:
\begin{itemize}
\item We find that the feature space of pre-trained VGG network is also effective for text-based style transfer.

% \item By utilizing the appropriate text style representation, we propose a new 
% text-based style transfer method called ITstyler.
% % feature space is optimized in the direction of style-consistent.
% In this work, we achieve a data-efficient text-based style transfer method that does not require optimization at the inference stage.

\item By utilizing the ability of VGG features to represent styles and the multi-modal embedding space of CLIP, we propose a data-efficient text-based style transfer method called ITstyler.

% \item Our method only need image data to train the network, solving the dilemma of requiring massive text-image paired data for arbitrary style transfer.

\item Our method can transfer arbitrary new styles of text input in real-time and synthesize high-quality artistic images.
\end{itemize}

\begin{figure*}[t]
    \centering
    \includegraphics[width=\linewidth]{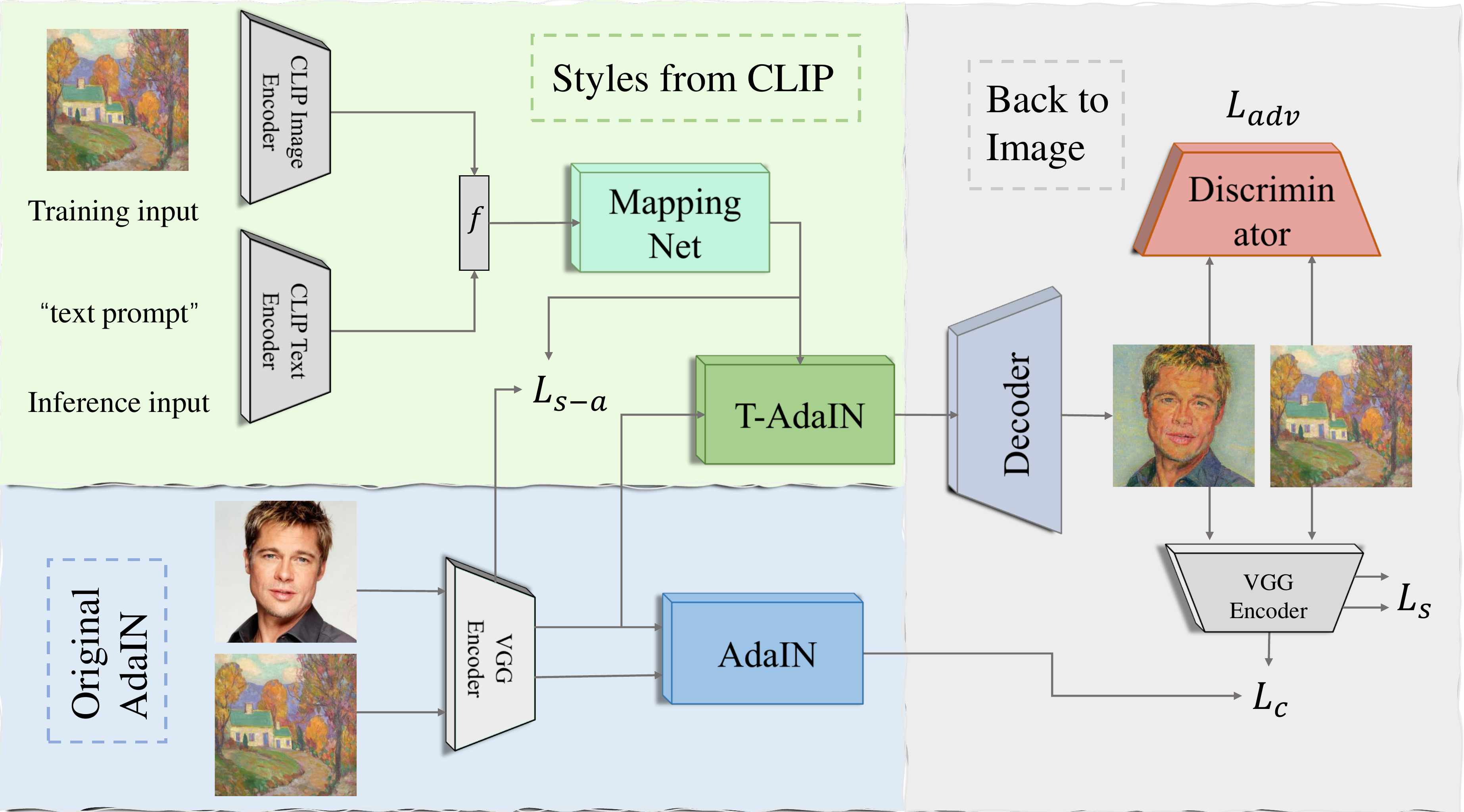}
    \caption{ The proposed ITstyler framework. We use CLIP's encoder to extract style embeddings $f_t^s(f_i^s)$ and then convert them to representations in VGG's style space $\mathcal{S}$ via a mapping network to achieve more effective style transfer in the space representing style well. During the training process, we only use image data. After the training, we can switch the input conditions to text with the help of CLIP's text-image shared feature space.}
    \label{fig:arch1}   
    % \vspace{-0.1in}
\end{figure*}

\section{Related Works}
\label{sec:related}
%写的时候可以参考其他文章的related work部分
\subsection{Style Transfer}

%先写之前的基于图像的风格迁移是怎么做的 随便列举一些工作（参考之前论文的related works），但是有几个比较关键的工作 Gatys et al. AdaIN，SANet注意要写
% DIN AdaAttn
Artistic style transfer aims to transfer the style from artworks to real-world photos and create images that have not appeared before.
% Artistic style transfer is aimed at transforming style patterns of images as well as preserving content structures. The seminal work of 
Gatys \etal. \cite{DBLP:conf/cvpr/GatysEB16} first introduced deep neural network to this field and proposed a neural style transfer method, which is based on an iterative pixel-optimization scheme. 
Then, AdaIN \cite{DBLP:conf/iccv/HuangB17} extended the style transfer method to synthesize the results in real-time by aligning the mean and variance between features of content and style.
% The AdaIN \cite{DBLP:conf/iccv/HuangB17} algorithm for the first time enables arbitrary style transfer in real-time by aligning the mean and variance between features of content and style. 
Further modifying the AdaIN, Jiang \etal. \cite{Jing_Liu_Ding_Wang_Ding_Song_Wen_2020} proposed the DIN module with lighter computational costs. SANet \cite{DBLP:conf/cvpr/ParkL19} proposed a learnable soft-attention-based machanism and introduces a identity loss, so that content features are matched with semantically nearest style features. Modifying SANet, Liu \etal. \cite{liu2021adaattn} proposed a adaptive attention normalization (AdaAttN) module for fast arbitrary style transfer. More recently, Kotovenko \etal. \cite{kotovenko_cvpr_2021} proposed a method that optimizes parameterized brushstrokes instead of pixel, which provide user control over the stylization process and better visual quality. Despite the inspiring results of previous works, reference style images have to be provided.

%然后再写最近的两个基于文本的是怎么做的以及他们的缺点
% Recently, several researches have attempted to deal with applications where only reference style texts are provided. embed style images and style instructions, extracting visual semantics of style instructions.
Recently, several studies have attempted to replace style images with texts that describe certain styles. By using CLIP, a pre-trained language-image embedding model, Kwon \etal. \cite {Kwon_2022_CVPR} proposed a patch-wise CLIP loss to align text-image pairs of source and target in the CLIP space. However, CLIPstyler trains a style-specific model for each target style, requiring extra time and resources. LDAST \cite{fu2022LDAST} learns to jointly extract style and content features from text and image data and fuse them to obtain results. However, LDAST has an obvious limitation of requiring text-image paired data for training. Overcoming the forementioned drawbacks, ITstyler supports arbitrary style transfer via global feature statistics matching and is trained using image data only.
% Per-Style-Per-Model(PSPM) trains separate style-specific models for each particular style, and is therefore burdensome to be adopted for real-world applications. 

\subsection{Text-driven Image Manipulation}
% 这里写基于文本的图像编辑 介绍下styleclip，hairclip，Tedi-GAN，StyleGAN-NADA
In addition to the works above, some text-driven image manipulation methods modify the properties of objects in the image based on input language. StyleCLIP\cite{Patashnik_2021_ICCV} utilized the state-of-the-art unconditional image generator StyleGAN\cite{karras2019style} and the powerful image text representation capability of CLIP\cite{radford2021learning} to modify images with given texts. It explores the learned latent space of StyleGAN and discovers appropriate editing directions towards text instructions. TediGAN\cite{xia2021tedigan} proposes a GAN inversion technique to map the image and text into a common StyleGAN latent space. However, both models are under limited practicality since they can merely be applied to do manipulation within trained image domain of the generator. StyleGAN-NADA\cite{gal2021stylegan} uses the semantic power of CLIP to guide the fine-tuning of a pre-trained generator to modulate existing generators from a given source domain to a new target domain, which solves the above-mentioned limitation to some extend. Nevertheless, it still heavily relies on pre-trained generative models. HairCLIP\cite{wei2022hairclip} achieves disentangled hair editing by feeding separate hairstyle and hair color information into different sub hair mappers to map the input conditions into corresponding latent code changes.

\section{Proposed Method}
\label{sec:method}

\subsection{Preliminary}

\textbf{AdaIN.} 
AdaIN \cite{DBLP:conf/iccv/HuangB17} presents an efficient solution for arbitrary style transfer and can render results in real-time.
It receives a content input $x$ and a style input $y$, and simply aligns the channel wise mean and variance of content feature maps to those of style feature maps as:
\begin{equation}
\mathbf{AdaIN}(x, y)=\sigma(y)\left(\frac{x-\mu(x)}{\sigma(x)}\right)+\mu(y).
\end{equation}
The input features of AdaIN are extracted from content image and style image by a pre-trained VGG-19 encoder, specifically the output of \verb+ReLU_4_1+ . After the style swap operation in the feature space, the output of AdaIN
is fed into a feed-forward decoder to get the final output. The reason why AdaIN can achieve arbitrary style transfer is that the mean of VGG features could encode brushstrokes in a certain style independent of content. Moreover, the variance of the features could encode more subtle style information, as explained in the original paper. In this work, we name the space of VGG features' mean and variance as Style Space $\mathcal{S}$, where the style representation of each image can be found.
%，每张图片的风格都对应着VGG特征的均值和方差，我们在这里将VGG特征的均值和方差的空间称为stylespace，每一张图片的风格都可以在这个空间找到它对应的风格表示
% 在该空间中都可应用其特征的来表示

\textbf{CLIP.}  
CLIP\cite{radford2021learning} is another recent advancement for language-image pre-training. It trains two encoders on an Internet-scale image set and learns a multi-modal embedding space connecting text and image features contains a wide range of visual concepts. The learned space has been found to be rich and effective for various downstream tasks. For example, as the space also encodes style-related features, it can be used for text-based style transfer. 
% In the later section, we call the CLIP's joint embedding space $\mathcal{X}$.
We name the subspace containing image styles of CLIP feature space as $\mathcal{X}$.

% learns a multi-modal embedding space, which
% may be used to estimate the semantic similarity between a
% given text and an image.

% In another recent advancement, CLIP [34], embedding
% spaces have been learned for image/text embedding on very
% large internet-scale data. The learned spaces, although
% trained using only loose image-keyword pairing information, have been shown to be rich and effective for several
% zero-shot tasks, i.e., not requiring any further training or
% fine-tuning for new tasks. For example, in the context of
% image manipulation, CLIP and StyleGAN have been utilized for text guided editing

% The CLIP model [28] is pretrained from 0.4 billion text-image pairs for cross112 modality semantic matching. This model consists of an image encoder and a text encoder to project
% an image and a text prompt into a 512-dimension embedding, respectively. We leverage the text
%  encoder of CLIP model to produce the text prompt embedding. This encoder is versatile in perceiving
%  text prompts with different semantic meanings.
% The CLIP image encoder [34] is trained on the common-crawl dataset, an
% internet-scale set of images that encompasses a broad range
% of visual concepts. 

\subsection{Styles in CLIP Space}
We assume that the image emdeddings obtained from the CLIP's encoder can also be divided into style $f_i^s$ and content $f_i^c$ parts in the CLIP emdedding space.
Therefore, CLIPstyler \cite {Kwon_2022_CVPR} can achieve text-based style transfer by utilizing CLIP's feature. It uses the text encoder to encode the text describing the style as $f_t^s$. Then, its stylization process can be regarded as using $f_t^s$ as a condition to guide $f_i^s$ to be consistent with the text.

Then we wonder: is it possible to directly swap style features in CLIP space to achieve more efficient text-based style transfer as previous image-based works did? On the one hand, it is difficult to decouple style and content from CLIP emdeddings. On the other hand, it is also hard to train a decoder to convert CLIP emdeddings back to images. However, we can adapt CLIP's subspace $\mathcal{X}$ to other spaces, such as the style space  $\mathcal{S}$ mentioned above, to achieve an effective style swap between text and image features.

\subsection{Adapt CLIP Space to Style Space}

Due to the lack of texts describing styles and the difficulty in finding the appropriate representation of the texts in $\mathcal{S}$, it is not an effective way to directly construct a mapping from text style emdeddings $f_t^s$ to $\mathcal{S}$ with text input. However, since the space of CLIP is shared by text and image features, we can learn such a mapping by using image data instead. This strategy has never been used before. The pipeline is shown in Figure \ref{fig:arch1}.

Following AdaIN, we use a pre-trained VGG \cite{DBLP:journals/corr/SimonyanZ14a} encoder to extract the features of the style images. The mean and variance of the \verb+ReLU_4_1+ layer are taken as the style representation in $\mathcal{S}$. At the same time, the style images are also sent to the CLIP image encoder to obtain 512-length emdeddings. Next, we use an MLP network to establish the mapping between the two representations:
\begin{equation}
    \begin{aligned}
        (\sigma_I,\mu_I) = \mathbf{MLP}(E_\mathcal{I}(I_s)),
    \end{aligned}
\end{equation}

where $I_s$ is the style image, and $E_\mathcal{I}$ is the CLIP's image encoder. After the training is finished, we can use this mapping network to find style representation of text emdedding $f_t^s$ in $\mathcal{S}$. 
Then we can use the following modified AdaIN operation to achieve arbitrary text-based image style transfer more efficiently:
\begin{equation}
\begin{split}
 (\sigma_T,\mu_T) = \mathbf{MLP}(&E_\mathcal{T}(T_s)),\\
    \mathbf{T\text{-}AdaIN}(x, \sigma_T,\mu_T)=\sigma_T&\left(\frac{x-\mu(x)}{\sigma(x)}\right)+\mu_T,
\end{split}
\end{equation}

where $(\sigma_T,\mu_T)$ is the style representation obtained from the text description and $x$ is the content input. The features transformed by T-AdaIN can be converted back to the image domain through a decoder to obtain the image with the style of text description.

\begin{figure*}[!ht]
    \centering
    \includegraphics[width=\linewidth]{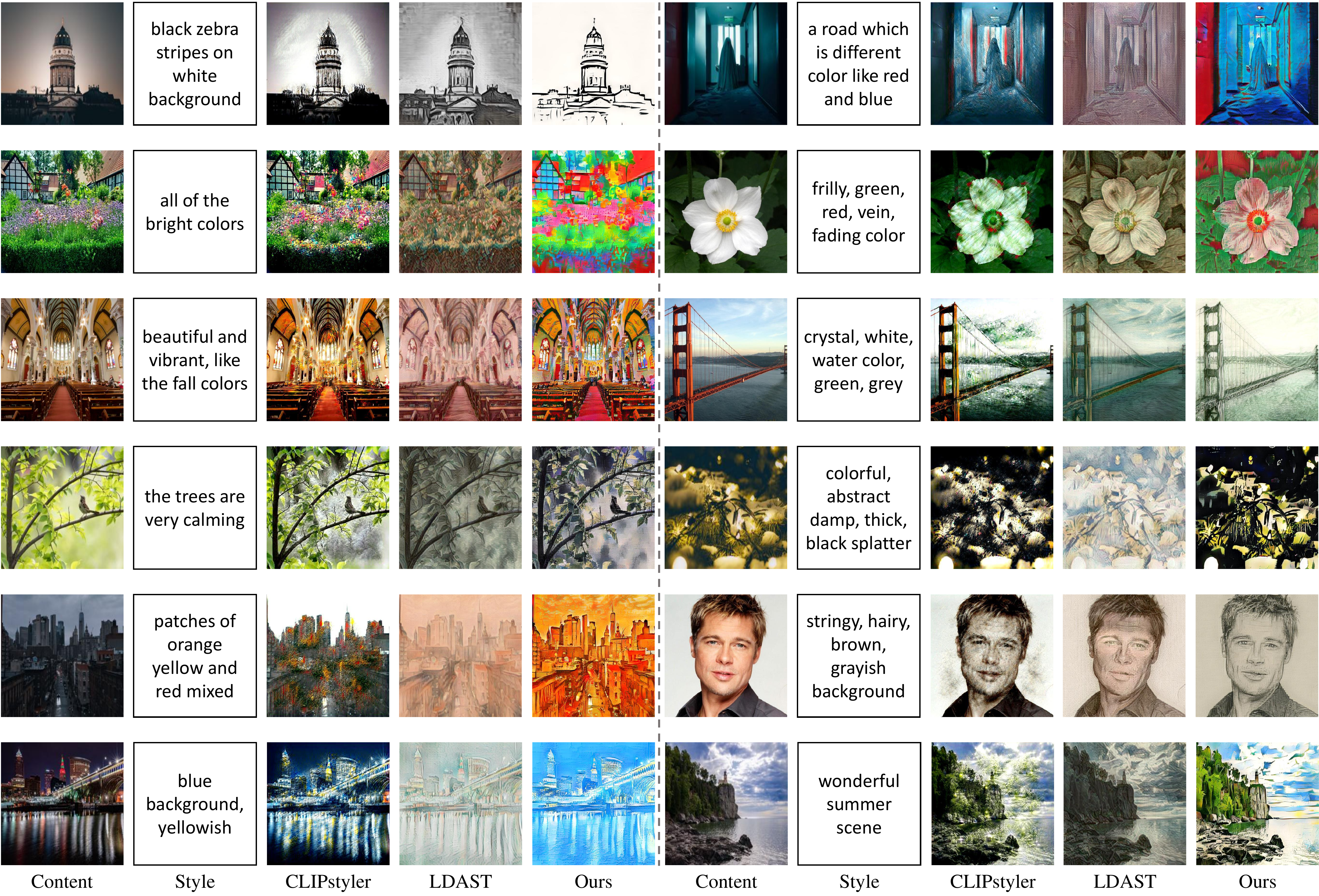}
    \caption{  Comparison results with previous text-based style transfer methods. The previous methods use inappropriate style representations that do not represent the style in the text well, resulting in poor style stylization. In contrast, the results of our method are more in line with the style described in the text and are more artistic and aesthetically pleasing.}
    \label{fig:com}   
    % \vspace{-0.1in}
\end{figure*}

\subsection{Objective Functions}

First, to train the mapping network, a style adaption loss is introduced between the output of the mapping network and the corresponding style representation:

\begin{equation}
\mathcal{L}_{s-a}=\left\|\mu_I - \mu\left(\phi(I_s)\right)\right\|_2+ \\
\left\|\sigma_I - \sigma\left(\phi(I_s)\right)\right\|_2,
\end{equation}

where $\phi$ is the \verb+ReLU_4_1+ layer of VGG-19. In the training process of the mapping network above, we also train a decoder $g$ to turn the feature $t = \operatorname{T-AdaIN}(f(I_c), \sigma_I,\mu_I)$ back to the image domain, generating the stylized image $T(I_c, I_s)$:

\begin{equation}
T(I_c, I_s)=g(t).
\end{equation}

Training this decoder requires similar content loss and style loss as original AdaIN:

\begin{equation}
\mathcal{L}_c=\|f(g(t))-t_o\|_2,
\end{equation}

where $t_o$ is obtained from the original AdaIN operation: $t_o = \operatorname{AdaIN}(f(I_c), f(I_s))$. The style loss is as follows:

\begin{equation}
\begin{array}{r}
\mathcal{L}_s=\sum_{i=1}^L\left\|\mu\left(\phi_i(g(t))\right)-\mu\left(\phi_i(I_s)\right)\right\|_2+ \\
\sum_{i=1}^L\left\|\sigma\left(\phi_i(g(t))\right)-\sigma\left(\phi_i(I_s)\right)\right\|_2,
\end{array}
\end{equation}

where each $\phi_i$ denotes a layer in VGG-19. We also use
\verb+ReLU_1_1+, \verb+ReLU_2_1+, \verb+ReLU_3_1+, \verb+ReLU_4_1+ layers with equal weights to compute the style loss.

In order to further improve the visual quality, we introduce an adversarial loss. All the above modules are regarded as a generator. We use a multi-scale discriminator to distinguish generated fake artworks from real ones.
The adversarial loss is as follows:

\begin{equation}
\mathcal{L}_{a d v}=\mathbb{E}\left[\log \left(\mathcal{D}\left(I_s\right)\right)\right]+
\mathbb{E}\left[\log \left(1-\mathcal{D}\left(T\left(I_c, I_s\right)\right)\right)\right].
\end{equation}

In summary, the overall loss function for the entire network is defined as: 
% Formula \ref{equ:total}:
\begin{equation}
\begin{aligned}
    \mathcal{L}_{total}&=\lambda_{s-a}\mathcal{L}_{s-a}+\lambda_{c}\mathcal{L}_{c}+\lambda_{s}\mathcal{L}_{s}+\lambda_{adv}\mathcal{L}_{a d v} \ ,
\end{aligned}
\label{equ:total}
\end{equation} 

where $\lambda_{s-a}$, $\lambda_{c}$,  $\lambda_{s}$ and $\lambda_{adv}$ are hyperparameters to constrain different loss terms.

% 为了更进一步地提升视觉质量，我们引入了对抗的损失。
% 上述的所有模块看做是一个生成器，一个多尺度的判别器被用于将真实图片与生成图片分开。
% multi-scale discriminator proposed bywhile the discriminator will try to distinguish generated fake artworks from real ones.
% 训练这个网络需要像adain一样的content loss和style loss

% 我们在训练上面的的映射网络同时训练一个decoder将特征转回图像域

% 交换过风格的特征再通过一个decoder转换回图像域就可以获得具有文本描述风格的图像。
%然后可以通过修改后的AdaIN操作实现任意的文本驱动图像风格迁移以一种更高效的方式。

%由于缺少大量的描述风格的文本并且也很难准确找到文本所对应在S中的表示，所以直接构造一个文本风格特征到S的映射不是一个有效的方式。但是由于clip的空间是文本和图像共享的一个空间，所以我们可以使用图像的数据来学习这样的映射
%如原来的Adain的操作一样 我们用一个预训练好的VGGencoder来提取风格图片的特征并使用第四层的特征的均值和方差来作为其风格在S中的表示。我们同时将风格图像送进clip的图像编码器获得512长度的特征，接下来我们一个mlp网络来建立两种特征之间的映射
%当训练完成后，我们就可以使用文本的编码器的通过这个映射网络找到其在S中对应的表示，然后实现

% 我们是否像可以像之前的工作一样直接交换style的特征来实现更加有效率的文本风格迁移呢？
% 不幸的是，一方面在clip的特征上做风格和内容的解耦比较困难，很难训练一个decoder将clip的特征转换回图像。但是我们可以将clip的空间X适应到其他的空间上去，比如上文中所提到的S，这样我们可以在另外的空间实现特征的风格交换。

% 不能因为没有这样的操作
% 但可以适应到其他的空间上

%我们假设图片编码到clip的特征空间中获得的特征也可以分为风格和内容两部分
% 我们将图片输入到clip的图像encoder中获得的特征中包含与内容相关的特征和与风格相关的特征两部分${\mathcal{f}_i}^c$ 和${\mathcal{f}_i}^s$
% clipstyler使用clip的文本的encoder将描述风格的文本条件编码到X中，作为condition来guide x 和文本的风格一致 
% guide the content image to follow the smantic of target text, the simplest CLIP-based image manipulation approac
% clipstyler使用
%图片的特征编码进来后我们的其中除了与风格有关的特征就是
% If IN normalizes the input to a single style specified by
% the affine parameters, is it possible to adapt it to arbitrarily given styles by using adaptive affine transformations? 

\section{Experiments}
\label{sec:experiments}

\subsection{Implementation Details}
% \vspace{-0.1in}
% we train our method using the COCO dataset [23] as content images and the WikiArt dataset [25] as style images.
The datasets for our experiments are the commonly used MS-COCO \cite{lin2014microsoft} (for the content images) and WikiArt \cite{karayev2013recognizing} (for the style images). Both datasets contain roughly $80,000$ training images. 
We use the Adam optimizer \cite{kingma2014adam} to train the mapping network and decoder for $160,000$ iterations, the learning rate is set to $1e\text{-}4$ and batch size is $8$. We set the coefficients for each loss function as follows: $\lambda_{s-a} = 10$, $\lambda_{c}= 5$, $\lambda_{s} = 10$, and $\lambda_{adv} = 1$. During training, the smaller dimension of the images is rescaled to $512$ while retaining the aspect ratio and then randomly cropped to the size of $256\times256$. We train our model on a single NVIDIA GeForce GTX 1080Ti GPU, and the training takes around $18$ hours. 

 The mapping network consists of $6$ fully-connected layers (using ReLU activations) that gradually convert the $512$-dimensional CLIP feature input into $1024$-dimensional style output. 
The decoder uses a mirror structure of the encoder without normalization layers, using nearest up-sampling layers to replace the pooling layers.
For CLIP, we follow CLIPstyler to use the ViT-B/32 model to extract text prompt embeddings.

% The optimizer (usually Adam \cite{kingma2014adam}) and the learning rate are the same as the corresponding methods.

%网络结构是怎么组成的
% the same as the corresponding methods. a batch size of 8 content-style image pairs. 
% with a learning rate of 0.0001 and a batch size of 16 for 160000 iterations.

\begin{figure*}[!ht]
    \centering
    \includegraphics[width=\linewidth]{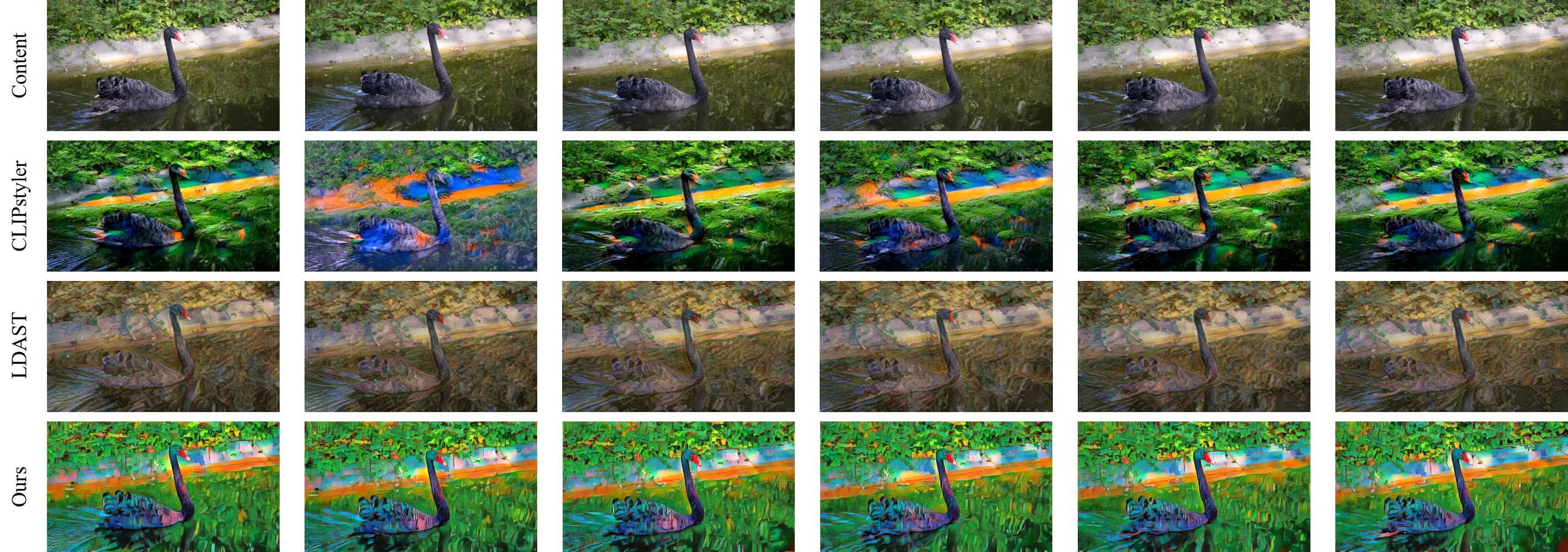}
    \caption{ Video style transfer. The video is stylized by this sentence: \emph{``The orange, blue, and green are all very light in color, very warm and inviting and make me want to go to this place.''}}
    \label{fig:video}   
    % \vspace{-0.1in}
\end{figure*}

\subsection{Qualitative Evaluation} We first qualitatively compare our method with two previous similar text-based style transfer methods.
Both two methods are conducted with public codes and default configurations. The comparison results are shown in Figure \ref{fig:com}. Most of the results of LDAST are grayish and do not match the style of the text description. This is because the attention mechanism it uses between the features of the two modalities is not accurate enough. 
As in the first and last rows of Figure \ref{fig:com}, the text contains the description ``blue,'' but the blue color is not presented in the LDAST results.
In some cases, CLIPstyler only makes very little stylization to the content image and adds some strange textures, resulting in unpleasant visual experience. For example, CLIPstyler's results look weird for ``golden gate'' in the third row and ``scenery waterside'' in the last row, illustrateing that the clip loss they use is not effective for style transfer. In contrast, the results of our method are more in line with the style described in the text, including some abstract words such as ``calming'' and ``wonderful summer.'' Moreover, the results generated by our method are more artistic and aesthetically pleasing.
%clipstyler在大多数情况下只是对内容图图原做了很少的风格化，并且是在原图上添加了一些奇怪的纹理，导致最终的结果不是很美观。
%相比之下我们的结果一方面符合文本所描述的风格 包括其中一些抽象的的词汇比如“wonderful summer”。并且我们的方法生成的结果更加具有艺术美感
%多加几行图的对比

\subsection{Quantitative Evaluation}

\textbf{User study.} User study investigates users' preferences for results of different methods for more objective comparison, which is the most widely used evaluation metric in style transfer. We also conduct such an user evaluation to verify the proposed method's effectiveness. In this part, we randomly sample $50$ groups from the results of above methods. Eventually, all $3\times50$ stylized images are distributed anonymously and randomly to $30$ college participants. They were asked to observe the stylized results from different methods and choose the image that better matches the style description with a high-quality visual effect. Each participant was asked to complete all choices within 20 minutes. As shown in Table \ref{tab:user}, the results of our method are mostly preferred.

\begin{table}
    \small
    \centering 
    \caption{The user study scores for different methods. The higher the better. Among the three methods, our results have the highest preference.}
\begin{tabular}{l|l} 
\hline
Methods & Preference Score  \\
\hline \hline CLIPstyler & \makecell[c]{12.1\%}  \\
LDAST & \makecell[c]{4.6\%}   \\
Ours & \makecell[c]{82.7\%} \\
\hline
\end{tabular}
    \label{tab:user}
\end{table}

\begin{table}
    \small
    \centering 
    \caption{The average LPIPS distances for different methods. The lower the better. Our method guarantees stable stylization in video frames.}
\begin{tabular}{l|l} 
\hline
 &  LPIPS distance  \\
\hline \hline Input & \makecell[c]{0.211}  \\
CLIPstyler & \makecell[c]{0.484}  \\
LDAST & \makecell[c]{0.317}   \\
Ours & \makecell[c]{\textbf{0.286}} \\
\hline
\end{tabular}
    \label{tab:lpips}
\end{table}

% For fairness, the images with the same reference are shown simultaneously in a random order. All participants were asked to observe the images for no more than $5$ seconds and choose the image that better matches the reference. As shown in Figure \ref{fig::pie}, we show the percentage of votes for each method in the form of pie chart. It shows that images of our method are mostly preferred.

\textbf{LPIPS.} We also extend our method to video style transfer and use Learned Perceptual Image Patch Similarity (LPIPS) \cite{DBLP:conf/cvpr/ZhangIESW18} as a quantitative indicator to measure the stability and consistency of frames like the practice in \cite{chen2021artistic}. The test videos are taken from the DAVIS \cite{Perazzi2016} dataset. We compute the average perceptual distances between adjacent frames, a lower LPIPS value represents better stability. Table \ref{tab:lpips} shows that our method gets the best score among all methods. Figure \ref{fig:video} shows an example of the video style results. It can be seen that CLIPstyler does not guarantee coherence between frames when applied to video scenarios, and the results vary greatly between frames. Although the style of the LDAST's results is consistent between frames, it does not match the text description. In contrast, our method ensures consistency between frames and transfers style from the text correctly.

\textbf{Speed analysis.}
Since our method can directly map arbitrary text to style representations and perform effective style transfer operations, we can achieve high efficiency in image processing. To demonstrate our method's superiority in speed, we also compare the running time with previous methods. 
The comparison results are shown in Table \ref{tab:speed}, all methods are tested on a single 2080Ti GPU. Our algorithm runs at $47.4$ fps and $25.6$ fps for $256 \times 256$ and $512 \times 512$ images, respectively. CLIPstyler requires a long optimization time for each text-image pair input individually. Their fast transfer solution can only perform one style transfer and the processing efficiency is also low. The attention mechanism used by LDAST is time-consuming. Our method is nearly $3$ orders of magnitude faster than CLIPstyler, and about twice as fast as LDAST. 
%由于我们的方法可以直接将任意的文本映射为风格表示并进行有效的风格迁移操作，我们的可以达到与AdaIN同样的处理图像效率。为了证明我们的方法在速度上的优越性，我们也与之前的方法比较了运行的时间，其中clipstyler对每个单独的文本图像对都要单独地长时间的优化，他们的fast transfer的方案只能做一种风格的迁移而他们的inference的效率也很低，LDAST的使用的注意力机制比较耗时。对比的结果如表1中所示，我们的速度相比clipstyler快三个数量级，相比于LDAST也能够快一倍左右
% our algorithm runs at 56 and 15 FPS for 256 × 256 and 512 × 512 images respectively
Our method is already fully applicable to process arbitrary user-provided styles in real-time. In some scenarios (e.g., transferring the same style to a video clip), the style text only needs to be encoded once and can be stored to process the entire video frame, so the speed of our method can be further increased when the encoding of the style is excluded.

\begin{table}
    \small
    \centering 
    \caption{Speed comparison (in seconds) for $256 \times 256$ and $512 \times 512$ images. 
    Our method is much faster than methods limited to one style and also faster than the LDAST applicable to arbitrary styles. Our (fast) stands for runtime excluding the time of style encoding. All the results are obtained with a single 2080ti GPU and averaged over 40 images.}
\begin{tabular}{l|l|l|c} 
\hline
Methods & Time (256px) & Time (512px) &  Styles \\
\hline \hline CLIPstyler & \makecell[c]{43} & \makecell[c]{49}& $\infty$ \\
CLIPstyler(fast) & \makecell[c]{0.116}  & \makecell[c]{0.122} & 1 \\
LDAST & \makecell[c]{0.036}  & \makecell[c]{0.063}  & $\infty$ \\
Ours & \makecell[c]{0.021} & \makecell[c]{0.039} & $\infty$ \\
Ours(fast) & \makecell[c]{0.015} & \makecell[c]{0.027} & $\infty$ \\
\hline
\end{tabular}
    \label{tab:speed}
\end{table}

\subsection{Ablation Studies}

\begin{figure}[t]
    \centering
    \includegraphics[width=\linewidth]{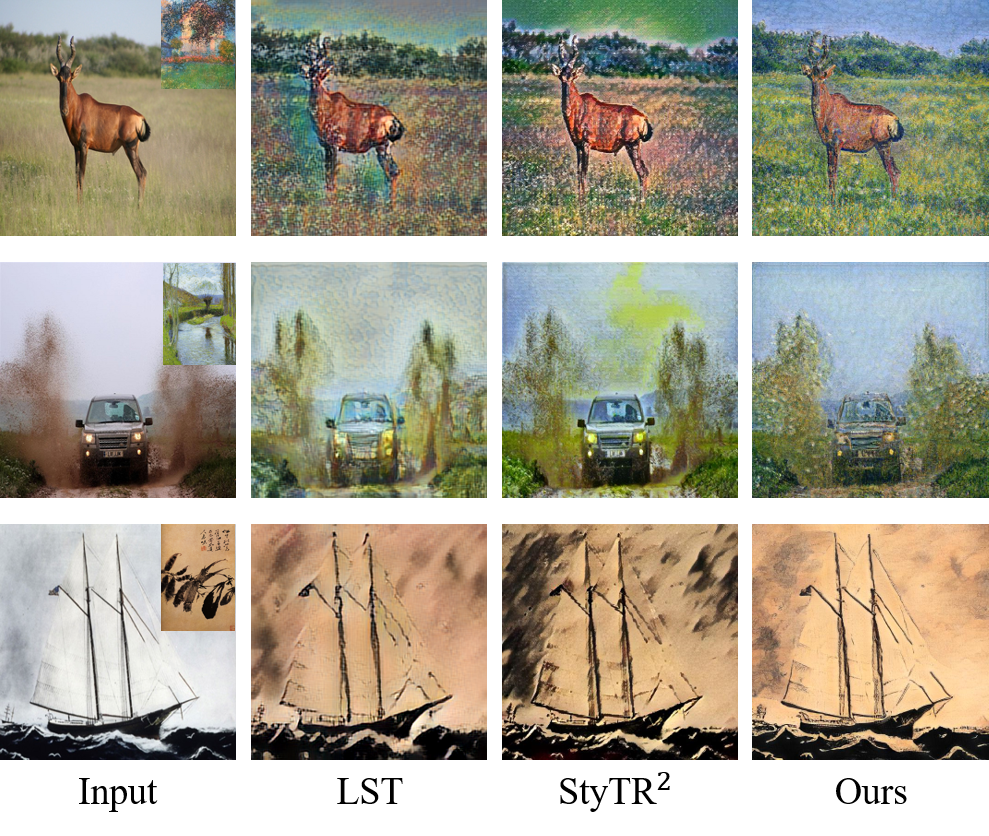}
    \caption{ Comparison with image-based methods. Our approach achieves similar or even better results. }
    \label{fig:image}   
    % \vspace{-0.1in}
\end{figure}

\textbf{Image-based style transfer.} By using image embeddings from CLIP to provide style, our approach naturally achieves image-based style transfer as well. We present the results of using images as style references here and compare them with two previous image-based methods. One is the widely used LST \cite{li2019learning}, and the other is the recent Transformer-based approach $\text{StyTR}^2$ \cite{deng2021stytr2}. As shown in Figure \ref{fig:image}, our approach achieves similar or even better results than the state-of-the-art image-based style transfer methods. These image-based methods sometimes generate color bleed artifacts, which are not present in our results. This also shows that our method has a wide range of application scenarios and high practical value.

\begin{figure}[t]
    \centering
    \includegraphics[width=\linewidth]{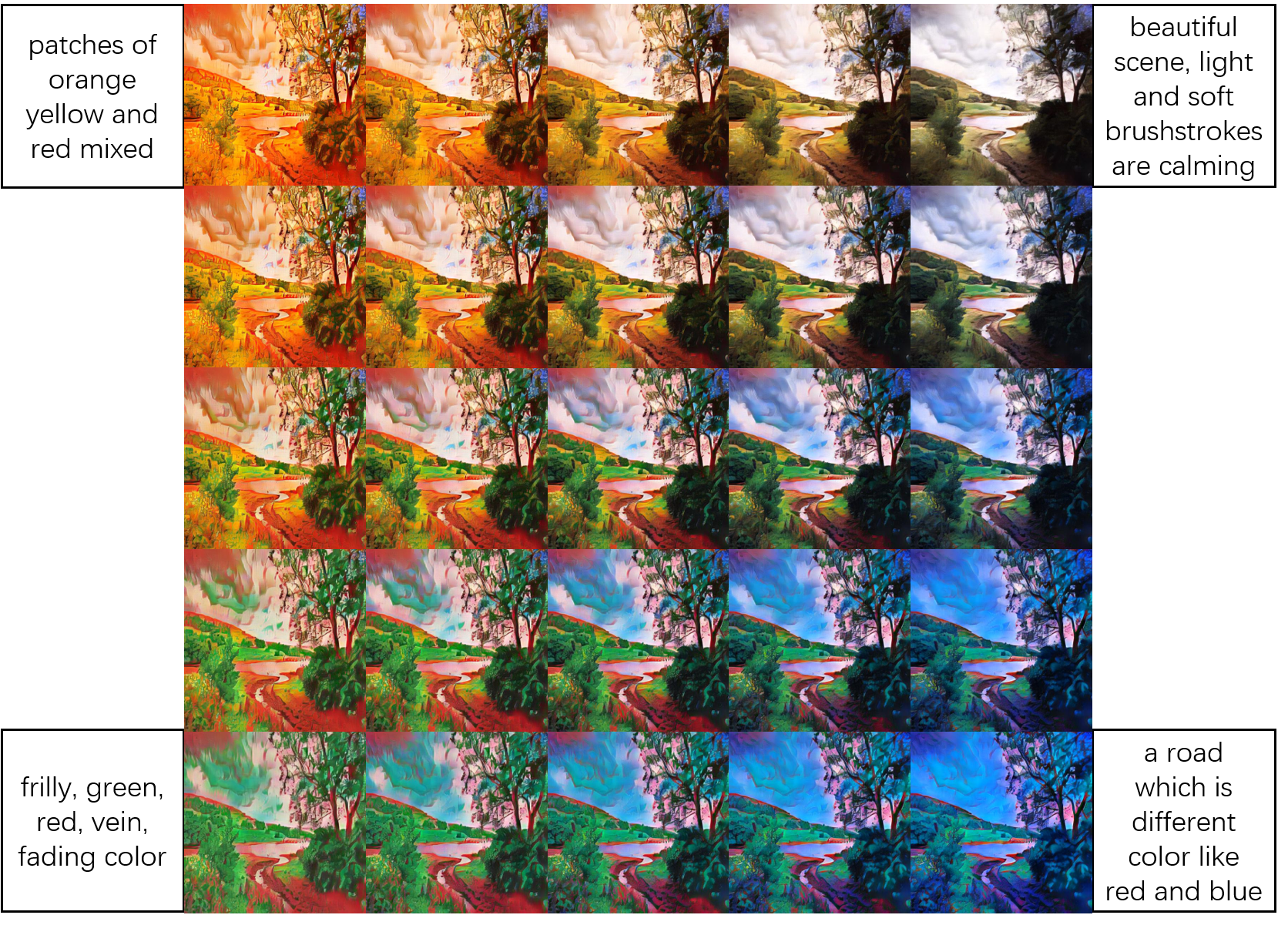}
    \caption{ Style interpolation example. By weighting the feature maps of different styles, we can smoothly interpolate between arbitrary style inputs to obtain a new style.}
    \label{fig:interpolation}   
    % \vspace{-0.1in}
\end{figure}

\textbf{Style interpolation.}
ITstyler can achieve smooth interpolation between arbitrary styles by weighting different feature maps with a set of weights $\{w_1, w_2, \dots, w_k\}$:
\begin{equation}
I=g(\sum_{k=1}^K w_k \mathbf{T\text{-}AdaIN}\left(f(c), \sigma_{T_k},\mu_{T_k}\right)),
\end{equation}

where $\sum_{k=1}^{K} w_k=1$. Figure \ref{fig:interpolation} shows an example of mixing four styles. The interpolation results are smooth and natural. Users can combine different styles in this way to get new styles.

%我们的方法也可以通过简单将特征进行加权来实现风格插值的效果
% To interpolate between a set of K style images s1, s2, ..., sK with corresponding weights w1, w2, ..., wK such that
% k=1 wk = 1, we similarly interpolate between feature maps (results shown in Figure \ref{fig:interpolation}):
% 用户可以通过这种方式将不同的风格进行组合获得新的风格

\begin{figure}[t]
    \centering
    \includegraphics[width=\linewidth]{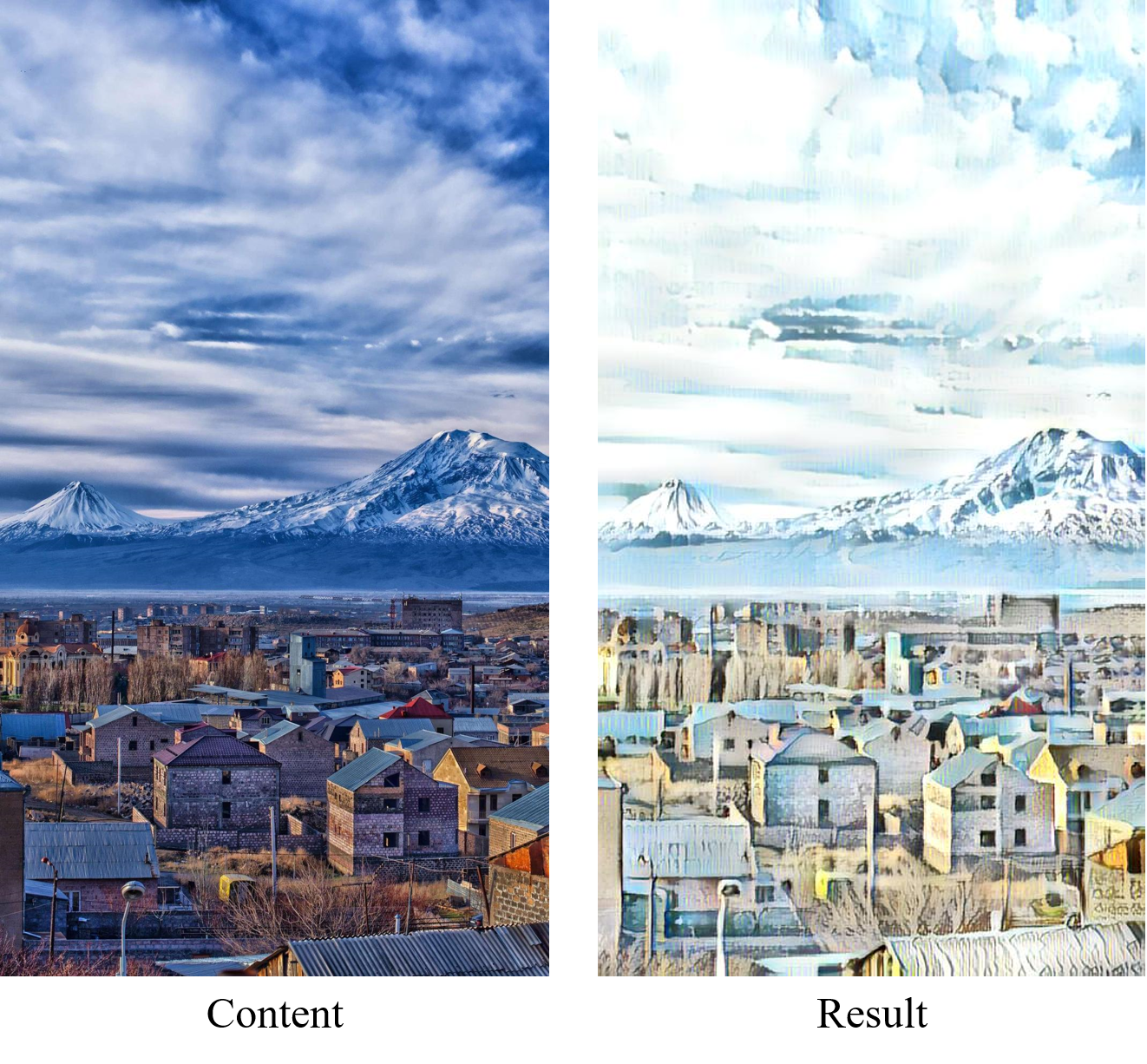}
    \caption{ High-resolution result. Our method can also perform style transfer on high-resolution images well. The rich details of the original image are also preserved. }
    \label{fig:High}   
    % \vspace{-0.1in}
\end{figure}

\textbf{High-resolution.} 
Since our network can process content images of various resolutions, we can also transfer styles to very high-resolution content images. In Figure \ref{fig:High},
we show the style transfer results for an image with a resolution of $1080 \times 1920$. It can be seen that our method achieves impressive stylization while preserving the rich details of the original image.

% Since our fast style transfer can be adapted to any kinds of content images with patch-based training, we can transfer the style with higher resolution content images. The training scheme is same as our fast style transfer method. After training the VGG model, we can feed the high-resolution image to trained network for style transfer. In Figure \ref{fig:High}, we can change the style of input to given text conditions while maintaining the details of the content.

%Our method can also perform style transfer on very high-resolution images, and the results have rich details
\begin{figure}[h]
    \centering
    \includegraphics[width=\linewidth]{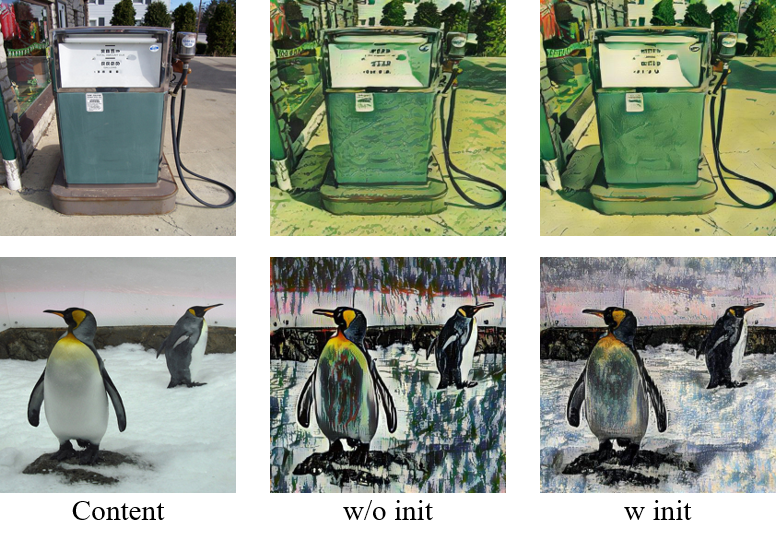}
    \caption{ Ablation study of the initial weights. The results generated by the network without initialization sometimes have strange strokes that affect the visual effect.}
    \label{fig:init}   
    % \vspace{-0.1in}
\end{figure}
\begin{figure}[t]
    \centering
    \includegraphics[width=\linewidth]{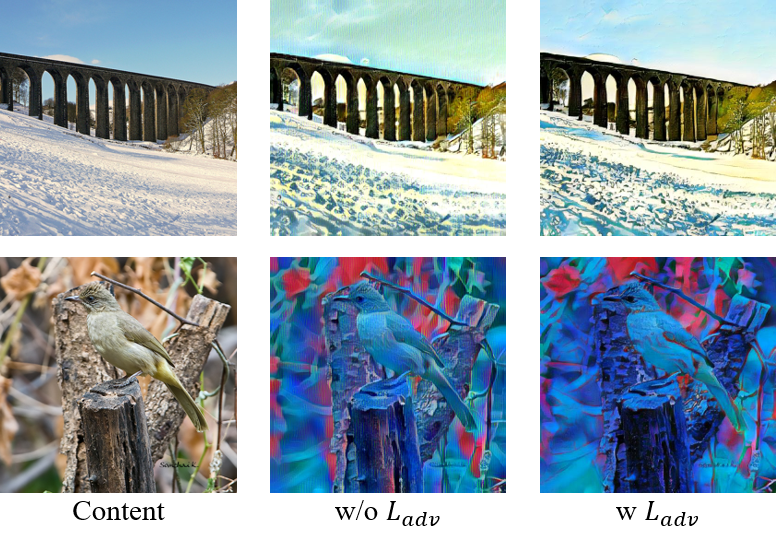}
    \caption{ Ablation study of the $L_{adv}$. The use of $L_{adv}$ improves the quality of the synthesized image and removes some grid artifacts}
    \label{fig:loss_adv}   
    % \vspace{-0.1in}
\end{figure}

\textbf{Training scheme.}
We find that initializing the parameters of the decoder with the pre-trained network parameters of the original AdaIN yields better results than random initialization. That is  because that AdaIN's decoder is already well-trained to convert the features back to stylized images. Figure \ref{fig:init} shows the comparison results with and without the use of initialization weights. The results generated by the network without initialization sometimes have strange strokes that affect the visual effect. In addition, in order to investigate the influence of the adversarial loss $L_{adv}$ used in the training process, we eliminate this loss and perform another ablation study. As shown in Figure \ref{fig:loss_adv}, the use of this loss improves the quality of the synthesized image and removes some grid artifacts.

\section{Conclusion}

In this work, we propose a new effective text-based style transfer method called ITstyler. Our method has two main advantages compared to previous text-based methods. First, our style swapping operation is performed in the feature space of pre-trained VGG. The space is more suitable for style transfer. Second, our method only requires image data to train the network, alleviating the dilemma of requiring text-image paired data.
Based on the above two points, our method is well capable of making the corresponding stylization based on the text input provided by the user. The most appealing property of ITstyler is that the training/testing phases use different data modalities. The results of our method are more artistic and aesthetic while matching the styles in the text description. Our method can get results in real-time and can be applied in various scenarios. Last but not least, our approach provides a new idea for linking text to appropriate style representations which will greatly facilitate subsequent related work.
% The most appealing property of ITstyler is that the training/testing phases use different data modalities. It adopts the image data for training but accepts the text data for testing. 
%在这个工作里我们提出了一个新的有效的基于文本的风格化方法叫做clipAdaIN。相比于之前的同类方法我们的方法主要有两个优势，一个是只需要图像数据来训练网络。另外的一个是我们的风格交换是在一个更加适合风格迁移的空间中进行的。%基于以上两点，我们的方法能够很好地根据用户提供的文本输入，来做对应的风格化。我们方法产生的结果在符合文本描述的条件的同时更加具有艺术美感。我们的方法可以实时得到结果，能够应用在更多场景中。

\begin{figure}[t]
    \centering
    \includegraphics[width=\linewidth]{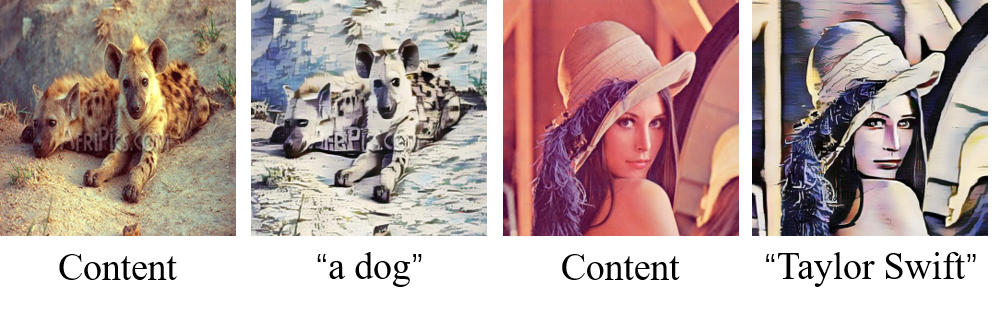}
    \caption{ Limitation. These two images are stylized by specific words input. For such text input, our method does not convert it into artistic style well.}
    \label{fig:Limitation}   
    % \vspace{-0.1in}
\end{figure}

\textbf{Limitation.}
Since our approach is an artistic style transfer method, the text input provided by users should describe a certain style. As shown in Figure \ref{fig:Limitation}, specific words such as ``a dog'' and ``Taylor Swift'' do not translate into a style representation well. However, this is not the scenario where style transfer is applied.
%由于我们的方法是做的艺术化的风格迁移，所以对于一些具体的词语输入比如“一条狗”，则无法很好地将其转为风格的表示. 也不能迁移一些真实的风格.

%%%%%%%%% REFERENCES
{\small
\bibliographystyle{ieee_fullname}
\bibliography{egbib}
}

\end{document}